\pdfoutput=1

\documentclass[11pt,table]{article}

\usepackage[preprint]{acl}

\usepackage{times}
\usepackage{latexsym}

\usepackage[T1]{fontenc}

\usepackage[utf8]{inputenc}

\usepackage{microtype}

\usepackage{inconsolata}

\usepackage{graphicx}
\usepackage{enumitem}
\usepackage{booktabs}
\usepackage{multirow}
\usepackage{amsmath}
\usepackage{xcolor}  
\usepackage{pgf} 
\usepackage{rotating}
\usepackage{subfigure}

\newcommand{\tf}{\textbf}

\usepackage{CJKutf8}

\title{From Tens of Hours to Tens of Thousands: \\Scaling Back-Translation for Speech Recognition}

\newcommand{\authcomma}{\textmd{,}\hskip0.5em}
\newcommand{\authspace}{\textcolor{white}{\authcomma}}
\author{
    Tianduo Wang$^{\heartsuit}$\thanks{Work done during internship at ByteDance} \authspace 
    Lu Xu$^{\diamondsuit}$ \authspace  
    Wei Lu$^{\heartsuit}$ \authspace 
    Shanbo Cheng$^{\diamondsuit}$\thanks{Corresponding author}\\
    $^\heartsuit$StatNLP Research Group, Singapore University of Technology and Design \\
    $^\diamondsuit$ByteDance Seed \\
    \texttt{\{tianduo\_wang,luwei\}@sutd.edu.sg} \\
    \authspace \texttt{\{xu.lu1,chengshanbo\}@bytedance.com}\\
    \url{https://github.com/tianduowang/speech-bt}
}

\begin{document}
\maketitle

\begin{abstract}
Recent advances in Automatic Speech Recognition (ASR) have been largely fueled by massive speech corpora. However, extending coverage to diverse languages with limited resources remains a formidable challenge.
This paper introduces Speech Back-Translation, a scalable pipeline that improves multilingual ASR models by converting large-scale text corpora into synthetic speech via off-the-shelf text-to-speech (TTS) models.
We demonstrate that just tens of hours of real transcribed speech can effectively train TTS models to generate synthetic speech at hundreds of times the original volume while maintaining high quality.
To evaluate synthetic speech quality, we develop an intelligibility-based assessment framework and establish clear thresholds for when synthetic data benefits ASR training. 
Using Speech Back-Translation, we generate more than 500,000 hours of synthetic speech in ten languages and continue pre-training Whisper-large-v3, achieving average transcription error reductions of over 30\%. These results highlight the scalability and effectiveness of Speech Back-Translation for enhancing multilingual ASR systems.
\end{abstract}

\section{Introduction}

Automatic Speech Recognition (ASR) technology has become increasingly important in making digital services accessible across languages and modalities~\cite{Baevski2020wav2vec2A,Zhang2021BigSSLET,Radford2022RobustSR}. While recent transformer-based architectures have achieved impressive results for high-resource languages, e.g., English and Chinese, many of the world's languages still lack sufficient transcribed speech for training robust ASR models~\cite{Pratap2020MassivelyMA,Babu2021XLSRSC,Chen2024TowardsRS}. This data scarcity creates a significant barrier to developing effective multilingual speech technologies, particularly affecting communities where manual data collection is resource-intensive or logistically challenging~\cite{team2022NoLL,Communication2023SeamlessME,Pratap2023ScalingST}.

A natural way to mitigate the data scarcity issue is to leverage high-quality generative models. Recent work has demonstrated successful applications of these models for data augmentation in computer vision~\cite{Fan2023ScalingLO, Azizi2023SyntheticDF}, natural language processing~\cite{Gunasekar2023TextbooksAA,Li2024SyntheticD}, and speech recognition~\cite{Yang2024EnhancingLA}. 
Despite their demonstrated potential, the role of generative models in overcoming data scarcity presents a paradox. 
These models themselves typically demand vast amounts of labeled data to attain their remarkable capabilities. For instance, Stable Diffusion~\cite{Rombach2021HighResolutionIS}, a leading text-to-image model frequently used for data augmentation~\cite{Tian2023StableRepSI,Trabucco2023EffectiveDA}, was trained on millions of labeled images.
This reliance prompts a fundamental question: do synthetic data truly alleviate data scarcity in downstream tasks, or do they simply shift the burden of data collection to the pre-training stage of generative models?

\begin{figure*}
\centering
  \includegraphics[width=0.8\linewidth]{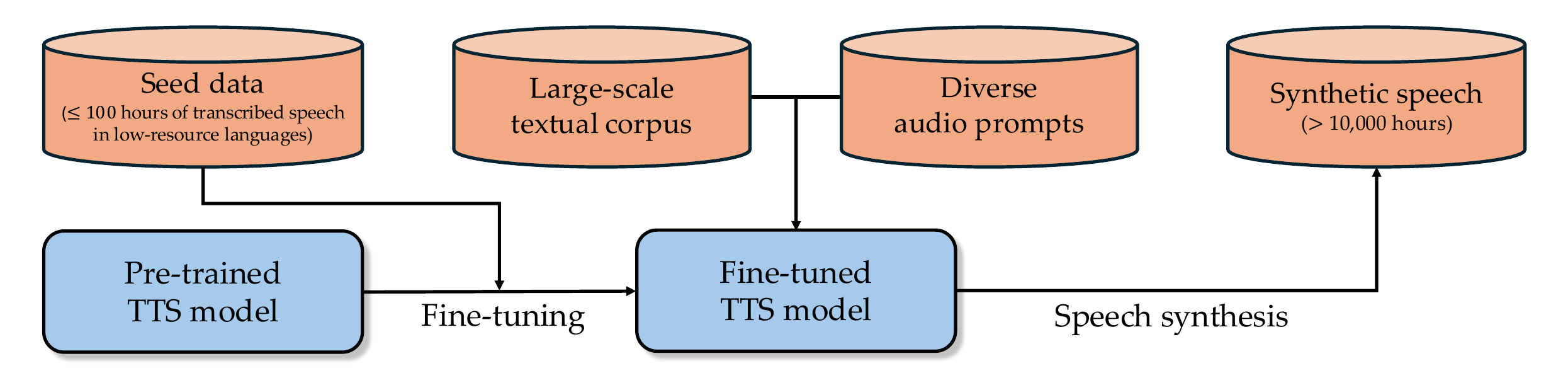}
  \vspace{-5pt}
  \caption{
    Pipeline of \tf{Speech Back-Translation}. 
    The main objective is to augment limited training data ($\leq$100 hours) for low-resource languages by synthesizing extensive amounts of speech ($>$10,000 hours). Starting from a  multilingual TTS model pre-trained with high-resource languages, we fine-tune it on a small set of seed data, then generate synthetic speech by conditioning the fine-tuned model on a large textual corpus and diverse audio prompts.
   }
  \label{fig:overview}
\end{figure*}

Our work investigates whether an off-the-shelf text-to-speech (TTS) model can be trained with limited real transcribed speech data—just tens of hours—to generate synthetic data that enhances multilingual ASR models. 
To address this challenge, we propose \tf{Speech Back-Translation} (see~\autoref{fig:overview}), a scalable method that build large-scale synthetic transcribed speech from text corpora with TTS models.
Our results demonstrate that synthetic data, generated by TTS models trained on just tens of hours of labeled audio, can effectively expand small human-labeled datasets to tens of thousands of hours.
To assess the quality of this back-translated synthetic dataset, we propose a novel intelligibility-based metric and use it to establish thresholds indicating when synthetic speech reliably enhances ASR performance.
Finally, we scale Speech Back-Translation to 500K hours across ten languages and continue pre-training Whisper-large-v3\footnote{\url{https://hf.co/openai/whisper-large-v3}}, one of the state-of-the-art multilingual ASR models. 
As a result, we observe consistent improvements across all languages, achieving an average reduction of over \tf{30\%} in transcription error rates.
To summarize, our main contributions are listed as follows:
\begin{enumerate}
    \item We demonstrate that just tens of hours of real transcribed speech can effectively train TTS models to generate tens of thousands of high-quality synthetic speech, achieving a scaling factor of several hundred.
    \item We introduce an intelligibility-based evaluation framework for synthetic speech and establish thresholds to determine when synthetic data reliably benefits ASR performance.
    \item We build the largest synthetic speech dataset to date—500K hours spanning ten languages—and use it to further pre-train Whisper-large-v3. This yields an average 30\% reduction on transcription error rates, highlighting the scalability of our approach.
\end{enumerate}

\section{Background}

\subsection{Back-Translation}

Back-translation is a data augmentation technique originally used in machine translation to expand training data~\cite{Sennrich2015ImprovingNM,edunov2018understanding}. 
In a typical setup, a model trained to translate from the target language back into the source language (i.e., a “{\em reverse}” model) is used to generate synthetic source sentences from real target-language data. 
These newly created source-target pairs can then be used to train a forward translation model, effectively increasing its exposure to a broader range of textual content. 
For speech recognition, back-translation offers a mechanism to supplement scarce or imbalanced datasets by leveraging an abundance of target-side text. 
Here, the “{\em reverse}” model is typically a text-to-speech model that generates synthetic speech from textual corpora. Integrating this synthetic speech with existing training data allows the model to handle a wider range of speech variability, enhancing recognition performance despite resource constraints.

\subsection{Zero-shot Text-to-Speech Model}

Zero-shot Text-to-Speech (TTS) models \cite{Wang2023NeuralCL, Casanova2024XTTSAM} represent a milestone in speech synthesis, enabling the generation of high-quality speech for previously unseen speakers without additional fine-tuning. 
These models typically contain the following components: 
\begin{itemize}[leftmargin=10pt,itemsep=0pt] 
    \item \textbf{Audio Tokenizer}: Encodes raw acoustic inputs (e.g., mel-spectrograms) into discrete audio tokens, forming the basis for synthesis. 
    \item \textbf{Speaker Embeddings}: Contain speaker-specific acoustic features, which are normally extracted from audio clips, enabling zero-shot adaptation to new voices.
    \item \textbf{Decoder-only Transformer}: Processes speaker embeddings alongside textual tokens to generate sequences of audio tokens. The Transformer model is trained in an auto-regressive manner.  
    \item \textbf{Vocoder}: Converts the generated audio tokens into waveform audio, producing the final synthesized output. 
\end{itemize}
The synergy of these components allows zero-shot TTS models to generalize effectively to speakers not encountered during training, maintaining high voice similarity and naturalness.

\section{Approach: Speech Back-Translation}

In this section, we introduce the proposed Speech Back-Translation (see~\autoref{fig:overview}).
we first detail how we extend existing TTS models to support new low-resource languages with fine-tuning (Section~\ref{sec:finetune}). We then describe how we generate large-scale synthetic speech dataset (Section~\ref{sec:gen_syn}).

\subsection{Fine-tuning with Low-resource Languages}\label{sec:finetune}

Obtaining high-quality transcribed speech for low-resource languages poses a significant challenge for multilingual ASR training. To address this, we extend existing multilingual TTS models—originally trained on high-resource languages—to new, low-resource languages via targeted fine-tuning with limited data. 

\paragraph{Vocabulary Expansion}
Before fine-tuning, we expand the vocabulary of pre-trained TTS models to accommodate words not encountered during the initial training phase. We employ the Byte-Pair Encoding algorithm~\cite{Sennrich2015NeuralMT} on textual data from the target language, appending the newly derived subwords to the model's original vocabulary. This approach preserves the integrity of the existing vocabulary while enabling effective representation of new linguistic units.
\paragraph{Limited Data Fine-tuning}
Given the scarcity of transcribed speech data, we adopt a conservative fine-tuning strategy: we freeze modules responsible for low-level acoustic representations, such as the audio tokenizer and vocoder, while selectively fine-tuning only the transformer part of the TTS model. This ensures stability in fundamental acoustic modeling while effectively adapting linguistic and prosodic mappings to the target language.
During fine-tuning, each pair of audio and transcript data is processed by first extracting a speaker embedding $\mathbf{e}$ from the audio clip. Then, we tokenize both the transcript and audio clip, concatenating the $S$ text tokens $\mathbf{x} = [x_1, \dots, x_{S}]$ and $T$ audio tokens $\mathbf{y} = [y_1, \dots, y_{T}]$ into $\mathbf{z} = [z_1, \dots, z_{S+T}]$. The training objective minimizes the negative log-likelihood of sequence $\mathbf{z}$ conditioned on the speaker embedding $\mathbf{e}$:

\begin{equation}
\mathcal{L} 
= -\sum_{t=1}^{S+T} \log \, p \bigl(z_t \mid z_1, \dots, z_{t-1}, \mathbf{e}\bigr),
\end{equation}

\paragraph{Quality Estimation}

Evaluating the performance of fine-tuned models is essential before deploying them for large-scale synthetic data generation.
Intelligibility—commonly measured as the Word Error Rate (WER) using a robust ASR system—has emerged as the standard metric for assessing synthetic speech quality~\cite{Wang2023NeuralCL,Casanova2024XTTSAM}. 
Yet this conventional method has two drawbacks:  
(1) the judge ASR introduces its own errors, particularly in low‑resource languages; and  
(2) absolute WER values are not comparable across languages.
To alleviate these issues, we propose a novel metric called {\em Normalized Intelligibility}, leveraging ASR performance on natural speech as a reference baseline. 
We use the Fleurs dataset~\cite{Conneau2022FLEURSFL}, which provides high-quality audio-transcript pairs across 102 languages, and Whisper-large-v3 as our judge ASR system.
By synthesizing speech using transcripts from Fleurs, we measure two WER scores for each language: WER on synthetic speech ($\text{WER}_s$) and WER on real speech ($\text{WER}_r$).
Normalized Intelligibility (Norm\_I) is defined as:
\begin{equation}\label{for:norm_i}
\text{Norm\_I} = \exp\left(\frac{\mathrm{WER}_r - \mathrm{WER}_s}{\mathrm{WER}_r}\right)
\end{equation}
This formulation offers several advantages: (1)  it normalizes ASR performance across languages using real speech as a baseline, (2) it enables meaningful cross-language comparisons, and (3) it produces intuitive scores bounded between 0 and $e$, where higher values reflect better synthetic speech quality relative to natural speech.

\subsection{Generating Large-scale Synthetic Speech}\label{sec:gen_syn}

Zero‑shot TTS converts \emph{text} into \emph{audio} by conditioning on two indispensable inputs:  
(i) an {\em audio prompt} that specifies the target voice style and  
(ii) a {\em text sentence} that supplies the textual content.  
Both inputs must therefore be covered at scale and with maximal diversity.

\begin{itemize}[leftmargin=10pt,itemsep=0pt] 
    \item \tf{Audio Prompts}: We curate around 1 million short audio clips spanning diverse speakers and recording conditions. After strict de‑duplication to remove near‑identical voices, every retained clip can serve as a style prompt that the TTS model imitates. Details of data sources and filtering are provided in~\autoref{sec:audio_prompt_details}.
    \item \tf{Text Corpus}: To maximize linguistic variety, we sample sentences across various domains, following the data‑mixing practices of recent open‑source LLMs~\cite{Touvron2023LLaMA1,Wei2023SkyworkAM}. Construction and statistics of the corpus appear in~\autoref{sec:text_data}.
\end{itemize}

\begin{figure}[t]
  \includegraphics[width=0.96\columnwidth]{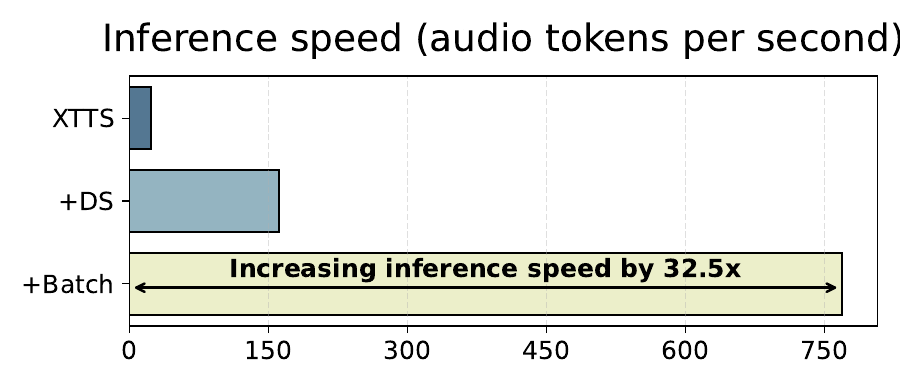}
  \centering
  \vspace{-10pt}
  \caption{
     \tf{XTTS inference speed} measured on a single NVIDIA V100-32GB GPU.
     ``DS'' refers to DeepSpeed-Inference while ``Batch'' refers to batch inference.
     For batch inference, we set batch size to be 16.
   }
  \label{fig:speed}
\end{figure}

\paragraph{Inference Speed-up}
A key challenge in employing TTS models for large-scale dataset creation is their inference speed. We address this bottleneck using two complementary optimization techniques:
\begin{itemize}[leftmargin=10pt,itemsep=0pt]
    \item {\em DeepSpeed‑Inference}~\cite{Aminabadi2022DeepSpeedIE}: Involving fused CUDA kernels integration and optimized kernel scheduling, significantly enhancing inference throughput.
    \item {\em Batch Inference}: We group multiple sentences with similar lengths using a single audio prompt, then apply tailored attention masks to enable simultaneous generation of multiple utterances in one forward pass.
\end{itemize}
We evaluate the effectiveness of these techniques using XTTS~\cite{Casanova2024XTTSAM} on a single NVIDIA V100 GPU.
As demonstrated in~\autoref{fig:speed}, we observe that these optimizations yield a more than \(\mathbf{30\times}\) speed‑up, making large‑scale speech synthesis feasible for our experiments. More details can be found in~\autoref{sec:app1}.

\section{Experimental Setup}

\subsection{ASR Backbone Models}
Our experiments leverage Whisper models~\cite{Radford2022RobustSR}, a family of multilingual ASR models pre-trained on 680,000 hours of labeled speech data, as the backbone.
The models are available in five sizes: Tiny (39M), Base (74M), Small (244M), Medium (769M), and Large (1.5B). Further training details are provided in~\autoref{app:training}.

\subsection{Zero-shot TTS Models}
We employ two state-of-the-art zero-shot TTS models in our experiments: XTTS~\cite{Casanova2024XTTSAM} and ChatTTS~\cite{chattts}. XTTS supports 16 languages, covering a range of language families and resource levels, 
while ChatTTS only supports Chinese and English. More details about these two models can be found in~\autoref{sec:comparative_study}.

\subsection{Languages}
Our experiments span ten languages across diverse language families and resource levels.
Following Whisper's training data distribution, we categorize them based on the relative resource availability:

\begin{itemize}[leftmargin=15pt, itemsep=3pt, parsep=3pt, topsep=6pt]
\item High ($\geq$10K hours):  English (\texttt{en}), Chinese (\texttt{zh}), French (\texttt{fr}), German (\texttt{de}), Spanish (\texttt{es})
\item Mid (1K$\sim$10K hours): Dutch (\texttt{nl}), Italian (\texttt{it})
\item Low ($\leq$1K hours): Vietnamese (\texttt{vi}), Czech (\texttt{cs}), Hungarian (\texttt{hu})
\end{itemize}

\noindent
Of these languages, XTTS supports all except Vietnamese. To enable Vietnamese support, we fine-tune XTTS with 100 hours of transcribed speech sampled from viVoice\footnote{\url{https://hf.co/datasets/capleaf/viVoice}}, a high-quality dataset derived from YouTube.

\subsection{Datasets}

Most of our experiments use Common Voice data~\cite{Ardila2019CommonVA}, chosen for its high quality and broad language coverage, and it also serves as the primary training corpus for XTTS. To assess generalization, we additionally evaluate our ASR models on Voxpopuli~\cite{Wang2021VoxPopuliAL} and Multilingual LibriSpeech~\cite{Pratap2020MLSAL}.

\begin{table}
  \centering
  \resizebox{0.8\linewidth}{!}{
  \begin{tabular}{lccc}
    \toprule
    \multirow{2}{*}{Model} & \multicolumn{3}{c}{WER$\downarrow$} \\
    \cmidrule(lr){2-4}
    & \texttt{vi} & \texttt{cs} & \texttt{hu}\\
    \midrule
    Whisper-medium   & 25.4  & 22.5  & 27.8       \\
    ~~~~+ Real-only  & 22.8  & 15.6  & 16.9      \\
    ~~~~+ Speech BT  & \tf{19.0} & \tf{10.3}  & \tf{13.2}       \\
    \midrule
    Whisper-large   & 24.5  & 19.9  & 23.8     \\
    ~~~~+ Real-only & 19.9  & 12.5  & 13.9      \\
    ~~~~+ Speech BT & \tf{16.0} & \tf{9.1}  & \tf{11.1}      \\
    \bottomrule
  \end{tabular}
  }
  \caption{
    \tf{WER results for low-resource languages on Common Voice}. The “Real-only” rows indicate models trained only on tens of hours of real audio, while the “Speech BT” rows present performance achieved when expanding training data to 10K hours using our method.
    }
  \label{tab:exp1}
\end{table}

\definecolor{darkgreen}{HTML}{006401}
\newcommand{\g}[2]{%
    \begingroup
    \ifdim -10 pt = 0 pt
      \pgfmathparse{50}%
    \else
      \pgfmathparse{round( (-#1)*100/(#2) )}%
    \fi
    \xdef\percentage{\pgfmathresult}%
    \cellcolor{darkgreen!\percentage} #1%
    \endgroup
}

\begin{table*}
  \centering
  \resizebox{0.8\linewidth}{!}{
  \begin{tabular}{lcccccccc}
    \toprule
    \multirow{2}{*}{\begin{tabular}[c]{@{}c@{}}\tf{Model}\end{tabular}} & \multicolumn{4}{c}{\textbf{Common Voice} (In-Domain)} & \multicolumn{4}{c}{\textbf{Voxpopuli} (Out-of-Domain)} \\
    \cmidrule(lr){2-5} \cmidrule(lr){6-9}
    & High   & Mid  & Low & Avg. $\Delta$ & High & Mid  & Low & Avg. $\Delta$\\
    \midrule
    Whisper-medium 
    & 11.5   &  10.6 &  25.2 & -          &  11.3 &  21.8 & 23.4  & -  \\
    ~~~~+ Real-only
    & 9.0   &  8.0 &  17.6 &\g{-4.0}{8}&  11.0 &  20.9 & 19.9 &\g{-1.4}{8}      \\
    ~~~~+ Speech BT
    & \tf{8.5}  &\tf{6.1}&\tf{11.1} &\g{-6.6}{8}&\tf{10.0}& \tf{19.4}& \tf{13.3}&\g{-4.1}{8}      \\
    \midrule
    
    Whisper-large 
    & 10.5  &  9.1   & 21.9  &-          & 11.4   & 20.3 & 18.1 & -      \\
    ~~~~+ Real-only          
    & 8.7  &  7.2    & 15.4  &\g{-5.0}{8}& 10.7   & 19.3 & 16.2 &\g{-1.2}{8}    \\
    ~~~~+ Speech BT
    & \tf{6.6}  &\tf{5.2} &\tf{10.7}&\g{-6.3}{8}&\tf{9.5} & \tf{17.7} & \tf{12.5}   &\g{-3.3}{8}   \\
    \bottomrule
  \end{tabular}
  }
  \caption{
  \tf{Comparison of Whisper models' WER across in-domain and out-of-domain data.} 
  Adding 3,800 hours of Common Voice data (Real-only) provides strong in-domain gains but limited out-of-domain improvements, whereas scaling synthetic Speech BT data to 160,000 hours achieves robust gains across both domains.}
  \label{tab:in-domain}
\end{table*}

\section{Results}\label{sec:pre_scale}

We begin by demonstrating the effectiveness of our approach in scaling limited real training data to tens of thousands of hours using synthetic speech (Section~\ref{sec:exp1}).
We then evaluate the models’ multilingual performance and examine their generalization to out-of-domain data (Section~\ref{sec:exp2}).
Next, we analyze the relationship between TTS quality and ASR performance using our fine-tuned TTS model (Section~\ref{sec:exp4}), and explore strategies for optimally leveraging limited in-domain real data (Section~\ref{sec:exp5}).
Finally, we scale the synthetic corpus to 500K hours and compare our results with prior work (Section~\ref{sec:exp6}).

\begin{figure}
\centering
  \includegraphics[width=0.95\columnwidth]{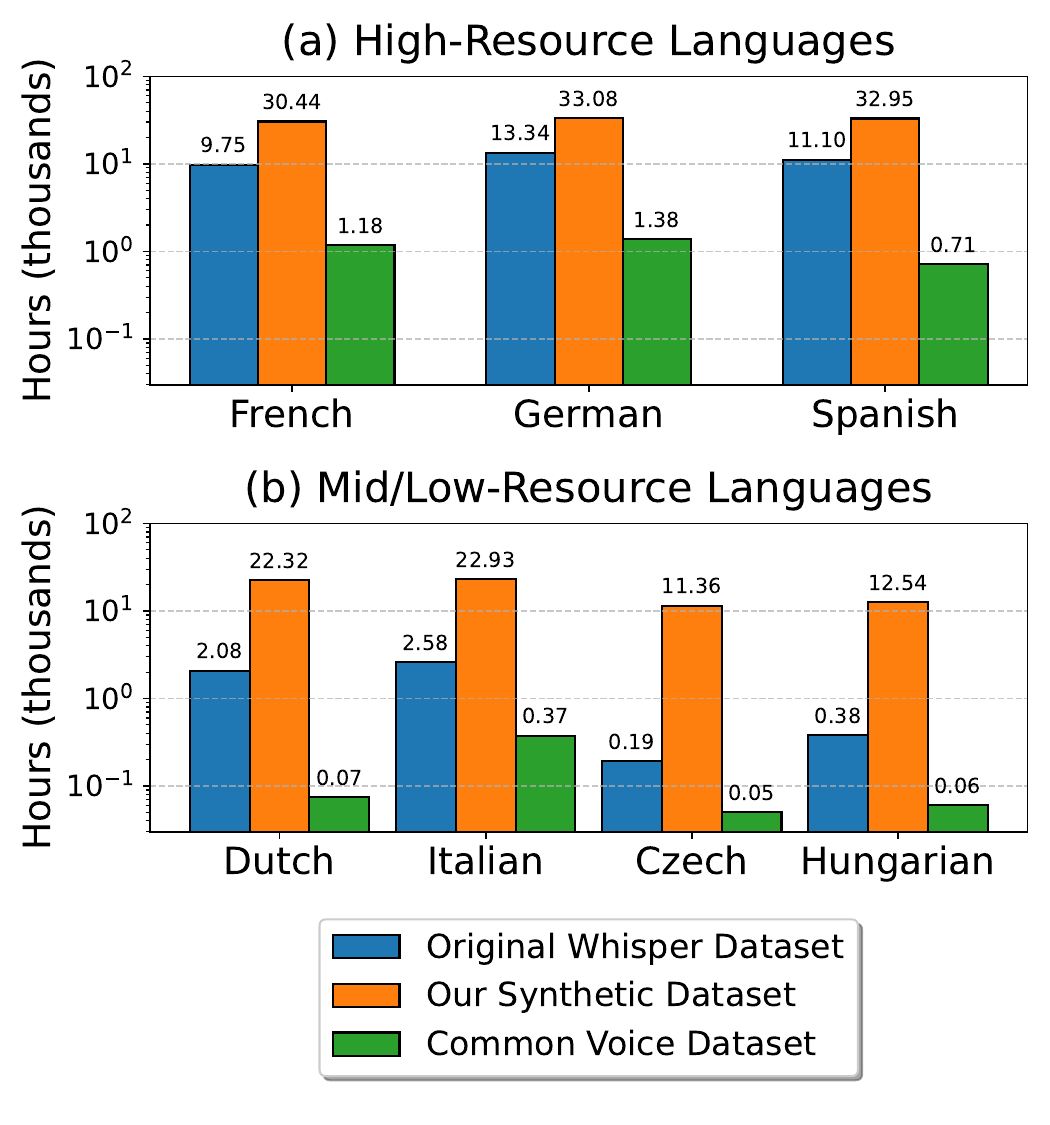}
  \vspace{-8pt}
  \caption{
    \tf{Comparison of dataset sizes across seven languages} (log-scale y-axis).
    Languages are categorized by resource availability in the Whisper dataset: (a) high-resource, (b) mid and low-resource groups.
   }
  \label{fig:160k}
\end{figure}

\subsection{From Tens of Hours to Tens of Thousands}\label{sec:exp1}
We first assess the effectiveness of our approach by expanding the amount of training data for three low‐resource languages—Vietnamese (\texttt{vi}), Czech (\texttt{cs}), and Hungarian (\texttt{hu})—from mere tens of hours to ten thousand hours.
As a baseline, we sample real audio in amounts matching the data originally used for TTS training (100 hours for \texttt{vi}, 50 hours for \texttt{cs}, and 60 hours for \texttt{hu})\footnote{Training data for \texttt{vi} comes from viVoice, whereas data for \texttt{cs} and \texttt{hu} are sampled from Common Voice.}.~\autoref{tab:exp1} compares these “Real‐only” models against models enhanced with our “Speech BT” method for both Whisper‐medium and Whisper‐large. Consistently across all three languages, Speech BT provides substantial gains in WER, underscoring the effectiveness of augmenting limited real speech with large‐scale synthetic data.

\subsection{Multilingual Performance and Out-of-Domain Generalization}\label{sec:exp2}
To evaluate the effectiveness and scalability of our approach in a multilingual setting, we generated 160,000 hours of synthetic speech spanning seven languages at varying resource levels: French, German, and Spanish (high-resource); Dutch and Italian (mid-resource); Czech and Hungarian (low-resource). As a baseline, we also collected 3,800 hours of transcribed speech from Common Voice as the training data.
\autoref{fig:160k} compares our synthetic dataset with the original Whisper Dataset and Common Voice. Our synthetic dataset provides substantially more training hours than the original Whisper dataset for each language: a 3-fold increase for high-resource languages, a 10-fold increase for mid-resource languages, and a 40-fold increase for low-resource languages.
While both Whisper training data and Common Voice exhibit substantial resource imbalance across languages (with high-resource languages having significantly more data than mid and low-resource ones), our Speech BT dataset maintains a more uniform distribution. This balanced allocation across language resources enables more equitable training, addressing a key limitation of naturally collected datasets.

\begin{figure}
  \centering
    \includegraphics[width=0.95\columnwidth]{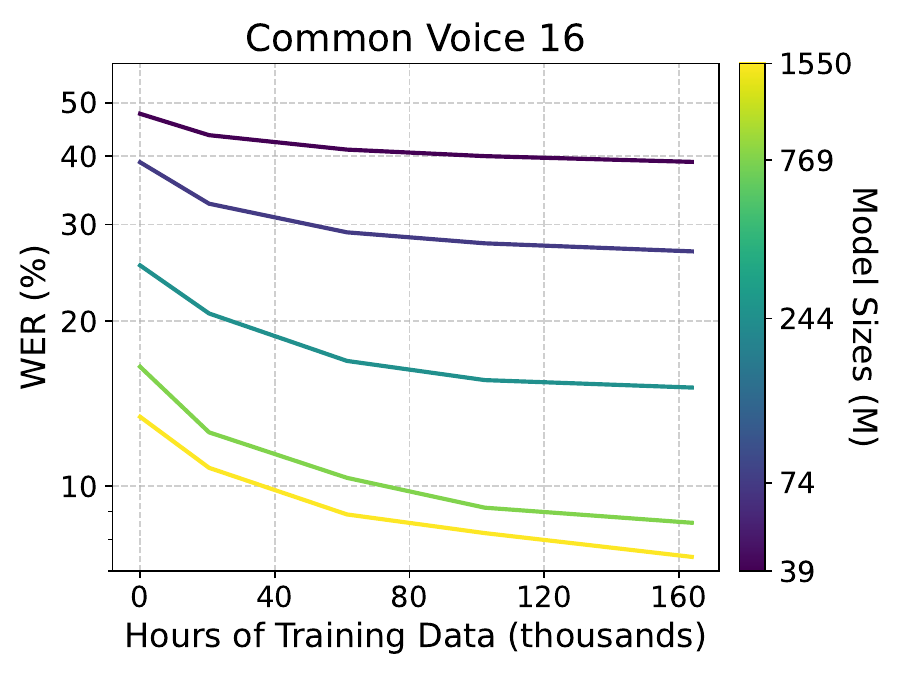}
  \caption{
    \tf{Whisper's performance improves consistently with larger models and more training data.}
    We train five sizes of Whisper models with up to 160,000 hours of data and conduct evaluation on Common Voice 16. We report averaged WER across seven languages.
  }
  \label{fig:scaling}
\end{figure}

\paragraph{Out-of-Domain Generalization}
\autoref{tab:in-domain} shows a detailed comparison of model performance in both in-domain (Common Voice) and out-of-domain (Voxpopuli) scenarios. Training with only real transcribed speech from Common Voice (Real-only) yields clear in-domain improvements for both Whisper-medium and Whisper-large (4.0\% and 5.0\% average WER reduction, respectively), but the generalization to out-of-domain data is limited (just 1.4\% and 1.2\% average reduction).
In contrast, supplementing real data with Speech BT significantly enhances both in-domain (6.6\% for Whisper-medium, 6.3\% for Whisper-large) and out-of-domain performance (4.1\% and 3.3\%, respectively). This clearly demonstrates that our synthetic data not only improves model robustness within-domain but also enhances generalization capabilities across diverse domains.

\paragraph{Scalability with Model and Data Size}
To further assess scalability, we train five Whisper model variants—tiny, base, small, medium, and large—using the same data mentioned above.
\autoref{fig:scaling} presents the averaged WER across all seven languages for each model size at increasing scales of training data up to 160,000 hours. The results show two clear trends. First, adding more training data consistently lowers WER across all model sizes.
Second, larger models achieve substantially lower WER at each data scale. These scaling trends suggest that our Speech BT approach effectively improves multilingual ASR performance across different model and data scales.

\begin{figure}[t]
\centering
  \includegraphics[width=0.9\columnwidth]{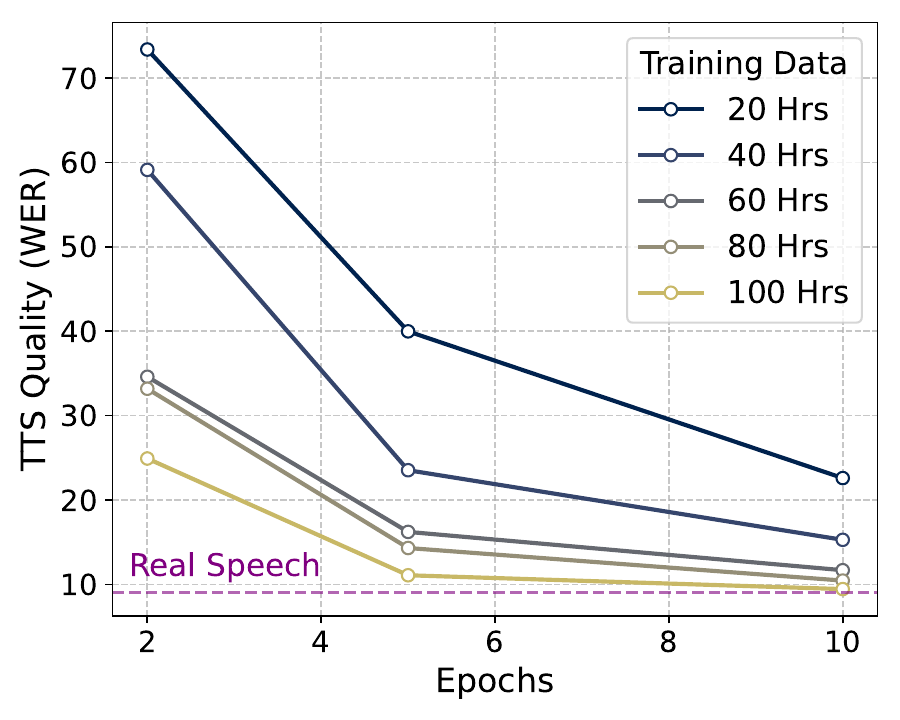}
  \caption{
    \tf{Impact of training data quantity and epochs on Vietnamese TTS quality.}
    The purple dashed line shows the WER of natural speech from Fleurs. 
  }
  \label{fig:viet_tts}
\end{figure}

\subsection{TTS Quality vs ASR Performance}~\label{sec:exp4}

\noindent
We now investigate extending TTS support to a new language, Vietnamese, using limited amounts of transcribed speech data. To explore how the quantity of training data impacts TTS model performance, we sampled datasets at increments of \{20,40,60,80,100\} hours and trained each for up to 10 epochs. The results, shown in~\autoref{fig:viet_tts}, clearly indicate that performance consistently improves as the amount of training data and the number of epochs increase. Specifically, the model trained on the 100-hour dataset reaching a WER of 10\% in the end, which closely approaches the baseline WER for natural speech.

\begin{figure}[t]
  \includegraphics[width=0.85\columnwidth]{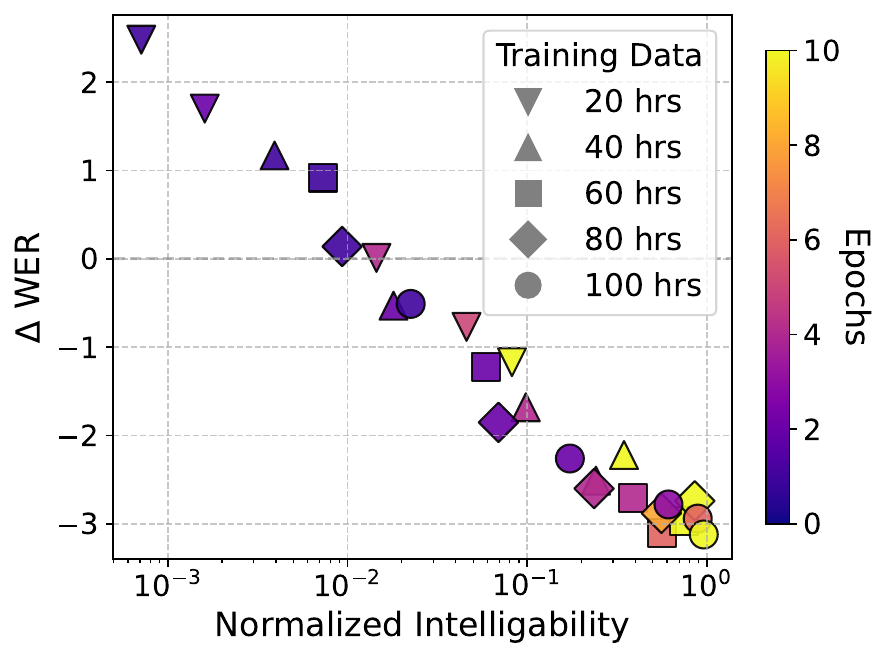}
  \centering
  \caption{
     \tf{Relationship between TTS quality and ASR performance.}
     Higher TTS intelligibility correlates with greater ASR improvement.
     }
  \label{fig:tts_asr}
\end{figure}

Next, we analyze the relationship between TTS model quality and downstream ASR performance. We selected several checkpoints from the fine-tuned TTS models, varying by the amount of training data and the number of epochs. For each checkpoint, we generated 100 hours of synthetic speech and subsequently used it to train Whisper-medium. We then measured the resulting changes in WER (denoted as $\Delta$WER, where negative values indicate improvement) on the Common Voice dataset. The correlation between each checkpoint’s normalized intelligibility score (see~\autoref{for:norm_i}) and ASR performance is illustrated in~\autoref{fig:tts_asr}.
Our analysis reveals a strong correlation between TTS intelligibility scores and ASR performance improvements. Notably, we identified a critical intelligibility threshold around 0.01, serving as a clear inflection point. Below this threshold, TTS-generated speech leads to increased WER, degrading ASR performance by up to 2 points. Conversely, once the threshold is surpassed, synthetic speech consistently enhances ASR accuracy, with greater intelligibility corresponding to more pronounced reductions in WER. This underscores the importance of achieving a minimum TTS quality level for effective ASR data augmentation.
Additionally, the volume of TTS training data significantly influences the ability to surpass this intelligibility threshold. Models trained on larger datasets generally achieve higher intelligibility scores and yield greater ASR performance gains. However, we observe diminishing returns as normalized intelligibility approaches 1.0, where WER reductions stabilize around 3 percentage points. This finding suggests that while adequate training data is essential to cross the quality threshold, further improvements in ASR performance may plateau beyond a certain point.

\subsection{Effective Utilization of Limited In-Domain Transcribed Audio}~\label{sec:exp5}

\noindent
The experiments in Section~\ref{sec:exp4} are essentially conducted under out-of-domain conditions, as the TTS models are trained on the viVoice dataset but the final ASR performance are evaluated with the Common Voice dataset. Notably, we identified only about three hours of transcribed audio in Common Voice Vietnamese available for training. This prompts an important research question: how can we effectively leverage such a small but valuable amount of in-domain data?
We propose three methods for effectively utilizing the in-domain data to enhance model performance:
\begin{itemize}[leftmargin=15pt, itemsep=3pt, parsep=3pt, topsep=6pt]
\item \textbf{Approach 1:} Pre-train Whisper on large-scale synthetic data followed by supervised fine-tuning using the limited in-domain data.
\item \textbf{Approach 2:} Prompt the fine-tuned Vietnamese TTS model with in-domain audio clips for speech synthesis.
\item \textbf{Approach 3:} Further fine-tune the Vietnamese TTS model with in-domain data before synthesizing speech.
\end{itemize}

\begin{table}[t]
  \centering
  \resizebox{0.64\linewidth}{!}{
  \begin{tabular}{lcc}
    \toprule
    Models & WER$\downarrow$  \\
    \midrule
    Whisper-medium     & 25.4         \\
    + in-domain fine-tune       & 21.6        \\
    \midrule
    \multicolumn{2}{l}{\it{Approach 1}}\\
    + Synthetic data pre-train    & 21.2        \\
    ~~~~+ in-domain fine-tune      & 20.4        \\
    \midrule
    \multicolumn{2}{l}{\it{Approach 2}}\\
    + Synthetic data pre-train    & 20.1        \\
    \midrule
    \multicolumn{2}{l}{\it{Approach 3}}\\
    + Synthetic data pre-train    & \tf{18.6}        \\
    \bottomrule
  \end{tabular}
  }
  \caption{Vietnamese WER performance on Common Voice using different approaches for leveraging limited in-domain data. }
  \label{tab:indomain}
\end{table}

\noindent
For all three approaches, we utilize fine-tuned XTTS checkpoints trained on 100 hours of transcribed speech and generate 1,000 hours of synthetic speech for pre-training Whisper-medium. 
\autoref{tab:indomain} summarizes the resulting WERs.
As a baseline, we first fine-tune Whisper-medium on three hours of in-domain data, which alone reduces WER from 25.4 to 21.6 and demonstrates the effectiveness of even limited domain adaptation.
By combining synthetic speech pre-training on 1,000 hours and subsequent in-domain fine-tuning (Approach~1), we obtain further WER reduction to 20.4.
However, the most pronounced improvements arise from leveraging in-domain data \emph{within} the TTS pipeline itself (Approach~2 and 3).
Simply prompting a fine-tuned XTTS model with in-domain audio achieves a WER of 20.1, already outperforming the synthetic pre-training baseline.
Adding a dedicated fine-tuning stage for XTTS on the three hours of Common Voice audio (Approach~3) yields the best overall WER of 18.6—a 27.0\% relative improvement over the 25.4 baseline. 
This underscores the value of adapting both the TTS and ASR models to the target domain, especially for low-resource languages like Vietnamese. 
In summary, leveraging a small in-domain dataset—through synthetic pre-training, language-specific fine-tuning, and TTS-based in-domain adaptation—proves highly effective for improving ASR performance under real-world low-resource conditions.

\begin{table*}[t]
  \centering
  \resizebox{0.95\linewidth}{!}{
  \begin{tabular}{lccccccccc}
    \toprule
    \multirow{2}{*}{\tf{Model}} & \multirow{2}{*}{\tf{Size}} & \multicolumn{3}{c}{\textbf{Common Voice}} & \multicolumn{3}{c}{\textbf{Voxpopuli}} & \multicolumn{2}{c}{\textbf{MLS}} \\
    \cmidrule(lr){3-5} \cmidrule(lr){6-8} \cmidrule(lr){9-10}
    & & High   & Mid  & Low  & High & Mid  & Low  & High & Mid  \\
    \midrule
    SeamlessM4T-medium &1.2B
    &13.3     &12.8     & 24.4     &10.7     &20.0    & 12.6 &8.0   &13.0   \\
    Whisper-large-v2 &1.5B
    &11.4     &8.1     & 19.9     &9.8     &19.5    & 16.3   &6.3    &11.5      \\
    \midrule
    Whisper-large-v3 &1.5B
    &10.1      &5.9    & 15.6    &12.6     &28.6     & 14.4  &5.3    &10.2    \\
    ~~~~+ Real-only (15K Hrs) & -
    &8.6      &4.9      & 12.5          &7.9      &17.1     & 10.6  &5.0    &9.4     \\
    ~~~~+ Speech-BT (500K Hrs) & -  
    &\tf{7.8} &\tf{4.3} &\tf{8.3} &\tf{7.6} &\tf{16.2} & \tf{8.0} &\tf{4.4} &\tf{7.6}   \\
    \bottomrule
  \end{tabular}
  }
  \caption{
  \tf{Multilingual ASR performance on various benchmarks.} 
  Results are averaged for each language resource category. 
  Word Error Rate (WER) is reported for all languages except Chinese, which is measured with Character Error Rate (CER).
  All results are normalized with Whisper Normalizer~\cite{Radford2022RobustSR}.}
  \label{tab:final}
\end{table*}

\subsection{Scaling to 500,000 Hours}~\label{sec:exp6}

\noindent
Building on insights from our previous analysis, we now push the limits of multilingual ASR training with Speech Back-Translation.
Starting from the baseline approach in Section~\ref{sec:exp2}, we implement several key enhancements:

\paragraph{Training Data Expansion}
We expand coverage to ten languages by incorporating three additional ones—English, Chinese, and Vietnamese. 
We also extend the amount of real speech: in addition to Common Voice~\cite{Ardila2019CommonVA}, we include real transcribed speech from Multilingual LibriSpeech~\cite{Pratap2020MLSAL}, Voxpopuli~\cite{Wang2021VoxPopuliAL}, and viVoice, bringing the total amount of real data to 15,000 hours.
Most significantly, we scale our synthetic speech dataset to 500,000 hours—a volume more than thirty times larger than the real data. The statistics of training data is illustrated in~\autoref{500k_data}.

\paragraph{Backbone Model and Baselines}
We adopt Whisper-large-v3, one of the state-of-the-art multilingual ASR models with 1.5B parameters, as our backbone model. For comparison, we include two ASR models with similar sizes—SeamlessM4T-medium~\cite{Communication2023SeamlessME} and Whisper-large-v2~\cite{Radford2022RobustSR}—as our baselines for their competitive performance and wide language coverage.

\paragraph{Results}
We evaluate both our models and baseline models on three benchmarks, Common Voice, Voxpopuli, and Multilingual LibriSpeech (MLS), and present the results in~\autoref{tab:final}.
We report the averaged results for each language category.
Results demonstrate a clear performance trajectory, 
training Whisper-large-v3 with 15K hours of real audio consistently improves performance across all benchmarks, while augmenting with 500K hours of Speech-BT data yields further substantial gains, achieving state-of-the-art results across all language categories.
On average across all benchmarks, our full model achieves a 30\% error rate reduction over the base Whisper-large-v3. Breaking this down by language groups, high-resource and mid-resource languages achieve 26\% and 30\% improvements respectively, while low-resource languages achieve a remarkable 46\% improvement.
These findings indicate that augmenting real data substantially with our synthetic Speech BT data contributes significantly to advancing multilingual ASR systems, with particular benefits for traditionally underserved language communities.
Detailed per-language results can be found in~\autoref{sec:additional_results}.

\section{Conclusion}
 
This work introduced Speech Back-Translation, a scalable approach to address the persistent challenge of data scarcity in multilingual ASR. 
Our method demonstrates that TTS models trained on merely tens of hours of transcribed speech can generate hundreds of times more synthetic data of sufficient quality to significantly improve ASR performance. 
The large-scale implementation across ten languages with 500,000 hours of synthetic speech yielded an average 30\% reduction in Whisper-large-v3's transcription error rates, confirming the effectiveness and scalability of our approach.
Speech Back-Translation challenges the need for massive human-labeled datasets by effectively scaling limited data, making advanced speech recognition more accessible across diverse languages. Future work could extend to extremely low-resource languages, refine language-specific metrics, and combine with other augmentation techniques.

\clearpage

\section*{Limitations}

While our approach demonstrates significant improvements in multilingual ASR performance, several limitations should be noted. 

First, the synthetic speech data generated through TTS models may not fully capture the acoustic complexity present in real-world environments, particularly in scenarios with background noise, multiple speakers, or variable recording conditions. This limitation could impact model robustness when deployed in settings with poor signal-to-noise ratios or challenging acoustic environments.

Second, although we introduce an intelligibility-based metric for assessing synthetic speech quality, this assessment framework may not comprehensively capture all relevant aspects of speech that could influence ASR training effectiveness. Future work could explore additional quality metrics that consider factors such as prosody and emotional expression.

Third, our experimental validation is primarily based on two TTS models (XTTS and ChatTTS), which may not represent the full spectrum of TTS capabilities and limitations. A more comprehensive evaluation across a broader range of TTS systems could provide additional insights into the generalizability of our approach and identify potential TTS-specific biases or artifacts.

Lastly, while we demonstrate the scalability of our method by generating 500,000 hours of synthetic speech, our language coverage remains limited to ten languages, with nine already supported by existing TTS models. Further research is needed to validate our approach's effectiveness in other low-resource languages, particularly those with distinct phonological characteristics or limited linguistic resources.

\section*{Acknowledgments}
This research/project is supported by the National Research Foundation, Singapore under its National Large Language Models Funding Initiative, (AISG Award No: AISG-NMLP-2024-005), and Ministry of Education, Singapore, under its Academic Research Fund (AcRF) Tier 2 Programme (MOE AcRF Tier 2 Award No. : MOE-T2EP20122-0011). Any opinions, findings and conclusions or recommendations expressed in this material are those of the authors and do not reflect the views of the National Research Foundation, Singapore, or Ministry of Education, Singapore.

\bibliography{custom}

\appendix

\section{Inference Optimization Details}\label{sec:app1}

We accelerate inference by integrating DeepSpeed-Inference~\cite{Aminabadi2022DeepSpeedIE} into the TTS pipeline. DeepSpeed’s \emph{deep fusion} merges multiple tiny CUDA launches into a single, highly optimized kernel that combines element-wise operations, matrix multiplications, transpositions, and reductions. Merging these operations reduce kernel-invocation overhead and off-chip memory traffic, translating into noticeably lower latency and higher throughput.
We compound these gains with \emph{batch inference}. Input sentences are grouped by language and length, then paired with a single audio prompt that supplies the target voice. Custom attention masks mark prompt–text boundaries, allowing the TTS model to synthesize multiple utterances concurrently. This batching strategy reduces redundant computations and GPU idle time, dramatically improving overall inference efficiency.

\section{XTTS vs ChatTTS}\label{sec:comparative_study}

In this section, we present a comparative analysis of XTTS and ChatTTS for generating synthetic audio in Chinese and English. 
~\autoref{tab:tts_models} summarizes the architectural details of both models.
As the XTTS's training data mainly come from Common Voice, we treat Common Voice 16 as the in‐domain dataset and Fleurs as the out‐of‐domain dataset for evaluation.

\begin{figure}
\centering
  \includegraphics[width=\columnwidth]{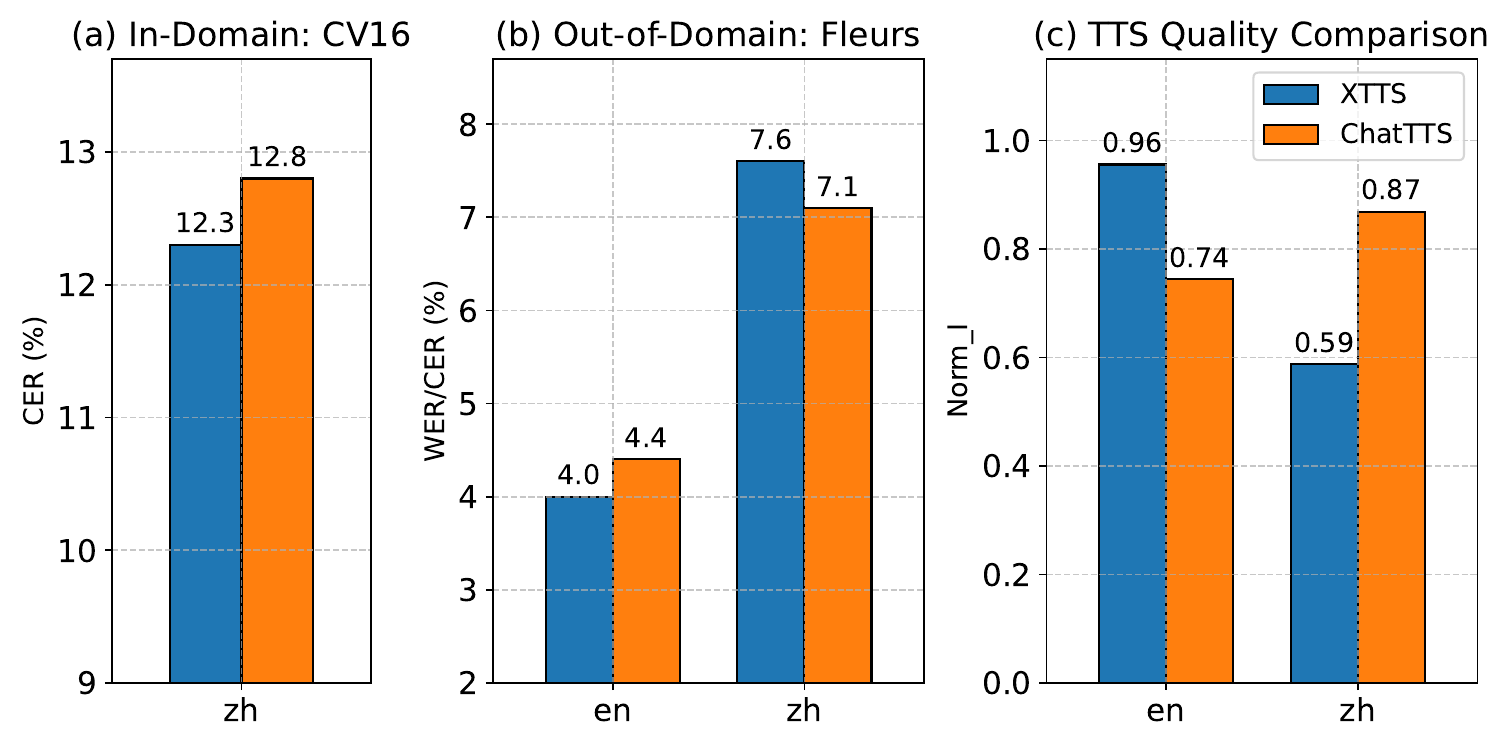}
  \caption{
    Comparison of Whisper-medium ASR performance on in‐domain (CV16) and out‐of‐domain (Fleurs) test sets, as well as TTS quality, when training with synthetic Chinese and English speech generated by XTTS and ChatTTS.
  }
  \label{fig:enzh}
\end{figure}

\paragraph{Performance Comparison}
We synthesize speech from 100K Chinese and English sentences using both models and train Whisper-medium to assess the effectiveness of these synthetic datasets.
As shown in~\autoref{fig:enzh} (a) and (b), XTTS outperforms ChatTTS on in-domain Chinese data, whereas ChatTTS excels on out-of-domain Chinese data. For English, XTTS achieves a WER of 4.0\%, surpassing ChatTTS’s 4.4\%. These trends highlight each model’s distinct strengths in handling language-specific characteristics.

\paragraph{TTS Quality Comparison}
To understand the performance difference, We assess the TTS quality with our proposed normalized intelligibility metric for both models.
As shown in Figure~\ref{fig:enzh}, XTTS achieves superior intelligibility in English (0.96 vs 0.74) while ChatTTS excels in Chinese (0.87 vs 0.59). 
Nevertheless, XTTS performs better on in-domain Chinese data, suggesting that while ChatTTS produces more intelligible speech in general, XTTS is more effective within the specific domain represented by Common Voice 16. This discrepancy may be attributed to domain‐matched acoustic patterns and speaking styles that XTTS models more accurately.
Meanwhile, on out-of-domain data (Fleurs), ChatTTS’s superior general intelligibility dominates, leading to stronger performance. In English, XTTS demonstrates higher intelligibility and more robust ASR results compared to ChatTTS. Overall, these findings underscore how a TTS model’s domain alignment and language‐specific strengths can influence synthetic data quality and downstream ASR performance.


\section{Audio Prompt Details}\label{sec:audio_prompt_details}

We collect a diverse set of audio clips from various sources to serve as audio prompts for our TTS models. To prevent redundancy in voice characteristics, we extract speaker embeddings from each reference clip using the ECAPA2 speaker encoder~\cite{Thienpondt2023ECAPA2AH} and remove duplicates by comparing their cosine similarity, applying a threshold of 0.8. Table~\ref{tab:audio_clip} summarizes the sources of these audio clips.

\begin{table*}
  \centering
  \resizebox{0.9\linewidth}{!}{
  \begin{tabular}{lcccccccc}
    \toprule
    \multirow{2}{*}{Model} & \multicolumn{3}{c}{Transformer} & \multicolumn{2}{c}{Vocabulary} & \multirow{2}{*}{Vocoder}& \multirow{2}{*}{Parameters} & \multirow{2}{*}{Lang}\\
    \cmidrule(lr){2-4} \cmidrule(lr){5-6}
    & Layers & Width & Heads & Text & Audio & \\
    \midrule
    {XTTS} & {30} & {1,024} & {16} & {6,681} & {1,024} & Hifi-GAN~\citeyearpar{Kong2020HiFiGANGA} & {467M} & {16}\\
    {ChatTTS} & {20} & {768} & {12} & {21,178} & {626} & Vocos~\citeyearpar{siuzdak2023vocos} & {280M} & {2}\\
    \bottomrule
  \end{tabular}
  }
  \caption{
    Architecture details of XTTS and ChatTTS. 
    }
  \label{tab:tts_models}
\end{table*}

\begin{table}
  \centering
  \resizebox{0.95\linewidth}{!}{
  \begin{tabular}{lcc}
    \toprule
    Dataset             & Num. Clips  \\
    \midrule
    Emilia~\cite{He2024EmiliaAE}              & 560K       \\
    CommonVoice~\cite{Ardila2019CommonVA}        & 230K       \\
    WenetSpeech~\cite{Zhang2021WENETSPEECHA1}         & 102K       \\
    CML-TTS~\cite{Oliveira2023CMLTTSAM}            & 92K       \\
    LibriTTS~\cite{Zen2019LibriTTSAC}             & 10K        \\
    \midrule
    Total           & 994K \\
    \bottomrule
  \end{tabular}
  }
  \caption{
    \textbf{Audio prompt distribution.} 
    The audio clips used for voice cloning comes from various sources.
    }
  \label{tab:audio_clip}
\end{table}

\section{Textual Data Details}\label{sec:text_data}

Our textual corpus is sourced from a wide range of domains. Since some sources include sequences that are too long for TTS synthesis, we first segment the text using a sentencizer. We then filter out sentences that are either too short, too long, or contain an excessive number of non-alphabetic characters. To reduce redundancy, we perform sentence-level de-duplication. A detailed breakdown of our corpus sources is provided below.

\paragraph{Wikipedia} Wikipedia is a collaborative online encyclopedia containing millions of articles, serves as a valuable source of high-quality natural text, therefore has been widely used for training language models~\cite{Touvron2023LLaMA1,Touvron2023Llama2}. 

\paragraph{WMT~\cite{wmt19}} We also collected textual data from the training split of WMT19 translation task, which is a widely-used training data source in machine translation research. 

\paragraph{Books} Our Books dataset is sourced primarily from Project Gutenberg, a digital library of public domain literature. Book-level de-duplication is performed to ensure the quality and uniqueness of the corpus.

\paragraph{Europarl~\cite{Koehn2005EuroparlAP}} Europarl is a parallel corpus created for training machine translation systems, containing aligned text in European languages extracted from European Parliament proceedings. We utilize 8th version of the dataset.

\paragraph{SkyPile~\cite{Wei2023SkyworkAM}} SkyPile is a large-scale Chinese dataset containing approximately 150 billion tokens, curated specifically for pre-training large language models. The corpus is compiled from diverse Chinese web pages across the public internet and undergoes rigorous quality control, including thorough document-level de-duplication and content filtering.

\section{Training Details}\label{app:training}

\paragraph{Whisper}
We train Whisper using AdamW ($\beta_1=0.9$, $\beta_2=0.98$, $\epsilon=1e-8$) with a weight decay of $0.01$. We use constant learning rate $7e-6$ after 5\% warm-up steps.
To optimize distributed training, we leverage DeepSpeed ZeRO-2~\cite{Rajbhandari2019ZeROMO}.
Additionally, we concatenate short audio clips—up to Whisper’s 30‐second input limit—to improve efficiency. Unless otherwise specified, our batch size is 128. In experiments presented in Section~\ref{sec:exp2}, we increase it to 768, while in Section~\ref{sec:exp6} experiment, we further increase it to 1,024. For evaluation, we generate transcripts with greedy decoding.

\paragraph{XTTS} 
Before fine-tuning, we expand the model's text vocabulary by incorporating 2,000 additional Vietnamese tokens by running Byte-Pair Encoding algorithms over Vietnamese textual data.
We used the AdamW optimizer $\beta_1=0.9$, $\beta_2=0.96$, and $\epsilon=1e-8$ with weight decay $0.01$, and a learning rate of 5e-6. The batch size is set to 32.

\section{Related Work}

\subsection{Synthetic Data for Multilingual ASR}
Recently we have witnessed the application of synthetic data in various domains and modalities, e.g., contrastive representation learning~\cite{Wang2022DifferentiableDA,Tian2023StableRepSI}, math reasoning~\cite{Wang2023LearningMR,Wang2024SelfTrainingWD}.
Our work focuses on improving multilingual ASR models using synthetic audio generated by zero-shot TTS models, with particular emphasis on low-resource languages. This research builds upon previous efforts that address data scarcity through synthetic data generation.
\citet{Bartelds2023MakingMO} demonstrated that both self-training and TTS-generated data can effectively overcome data availability limitations in resource-scarce languages. Their work specifically examined four languages: Gronings, West-Frisian, Besemah, and Nasa, showing significant improvements in ASR performance.
\citet{Baas2021VoiceCC} explored voice conversion (VC) models for data augmentation in low-resource languages. Their key finding was that a VC system trained on a well-resourced language like English could generate effective training data for previously unseen low-resource languages.
More recently, \citet{Gao2024SpeechSL} proposed using diffusion models to generate high-quality synthetic audio for self-supervised pre-training. The authors suggest that diffusion models are particularly adept at capturing complex speech structures from real audio, making the synthetic data especially valuable for self-supervised learning tasks.

\subsection{Text-Based Back-Translation}
Back-Translation~\cite{Sennrich2015ImprovingNM,edunov2018understanding} is originally proposed machine translation~\cite{Sennrich2015NeuralMT,Pan2024GDIGTG} to augment the limited parallel training corpus from the large amount of monolingual textual data.
It is designed to translate the target-language data into the source language, generatin additional synthetic parallel data that boosts overall translation quality~\cite{Sennrich2015ImprovingNM}.
This method capitalizes on monolingual text resources, which are more abundant than parallel corpora, thereby increasing model robustness and reducing overfitting.
Subsequent work has explored variants of back-translation such as iterative back-translation, filtering synthetic data by quality, and domain adaptation strategies~\cite{edunov2018understanding,Hoang2018IterativeBF}.
In addition, dual learning frameworks have incorporated back-translation and forward-translation jointly for unsupervised and semi-supervised machine translation scenarios~\cite{He2016DualLF,Lample2017UnsupervisedMT}.
These developments underscore the broader impact of synthetic data in enhancing model performance, even where labeled data are sparse.

\subsection{Speech Translation}
Beyond text-based machine translation, speech translation deals with converting audio signals in one language to either text or audio in another language, frequently via cascading automatic speech recognition and machine translation modules or through end-to-end systems~\cite{Cheng2024TowardsAH,Huang2023SpeechTW}.
One persistent challenge in this domain, especially for lower-resource languages, is the scarcity of paired audio-transcript data. A widely used approach to address this limitation is to create pseudo-labeled data by transcribing existing audio and then translating the resulting transcripts~\cite{Communication2023SeamlessME,puvvada2024less}. A natural future direction for our Speech Back-Translation approach could be extended to speech translation tasks by synthesizing speech from existing parallel corpora.

\section{500K-Hour Training Data Statistics}\label{500k_data}
The detailed statistics of training data used in our 500K-hour scaling up experiments are presented in~\autoref{tab:500k_data}. 

\begin{table}
  \centering
  \resizebox{0.75\linewidth}{!}{
  \begin{tabular}{lcc}
    \toprule
    \multirow{2}{*}{Language} & \multicolumn{2}{c}{Amount (Hrs)} \\
    \cmidrule(lr){2-3}
    & Real & Synthetic \\
    \midrule
    English  & 3,951  & 75,159   \\
    French   & 2,486  & 94,822   \\
    German   & 3,706  & 90,782   \\
    Spanish  & 1,674  & 47,745   \\
    Chinese  & 204    & 37,910   \\
    Dutch    & 1,525  & 41,095   \\
    Italian  & 839  & 38,069   \\
    Czech    & 119  & 33,312   \\
    Hungarian  & 156  & 33,492   \\
    Vietnamese & 104  & 13,444   \\
    \midrule
    Total    & 14,864  & 505,830 \\
    \bottomrule
  \end{tabular}
  }
  \caption{
    Statistics of the training data in our 500K-hour experiment.
    }
  \label{tab:500k_data}
\end{table}

\section{Additional 500K-Hour Scaling Results}\label{sec:additional_results}
In this section, we show detailed results for each languages from Section~\ref{sec:exp6}.
The results for Multilingual Librispeech (MLS), Voxpopuli, and Common Voice 16 are presented in Table~\ref{tab:whisper-c}, Table~\ref{tab:whisper-vox}, and Table~\ref{tab:whisper-cv} respectively.
Additionally, we make comparisons with state-of-the-art multilingual ASR models: SeedASR~\cite{Bai2024SeedASRUD}, SeamlessM4T~\cite{Communication2023SeamlessME}, Canary~\cite{puvvada2024less}, and Whisper-large and Whisper-large-v2~\cite{Radford2022RobustSR}.

\begin{table*}
\centering
\begin{tabular}{lr|cccccc}
\toprule    
\multirow{2}{*}{Model} & \multirow{2}{*}{Size} & \multicolumn{4}{c}{High} & \multicolumn{2}{c}{Mid} \\
\cmidrule(lr){3-6} \cmidrule(lr){7-8}
& & \texttt{en} & \texttt{fr} & \texttt{de} & \texttt{es} & \texttt{nl} & \texttt{it} \\
\midrule
SeedASR             & -      &  4.1   & 5.1  & -    & 3.8  &  -    &    - \\
SeamlessM4T-medium  & 1.2B   &  9.8   & 7.9  &  8.9 & 5.4  & 13.6  & 12.3\\
Canary              & 1.0B   &  5.1   & 4.4  &  4.7 & 3.4  &  -    & -\\
\midrule
Whisper-large               &1.5B  & 7.2 & 8.8 & 7.4& 5.3& 11.1  & 14.1 \\
Whisper-large-v2            &1.5B  & 6.8  & 7.4 & 6.4 & 4.6 & 10.0  & 12.9 \\
\midrule
Whisper-large-v3    &1.5B  & 5.3  & 5.6 & 6.0 & 4.0 & 10.4  & 9.9 \\
~~~~+ Real-only (15K Hrs)    & -    & 5.5  & 5.1 & 5.7 & 3.5 & 10.2  & 8.5 \\
~~~~+ Speech BT (500K Hrs)    & -    & 5.2  & 4.3 & 4.9 & 3.0 & 8.5  & 6.7 \\
\bottomrule
\end{tabular}
\caption{Performance comparison across languages on Multilingual LibriSpeech (MLS).}
\label{tab:whisper-c}
\end{table*}

\begin{table*}
\centering
\begin{tabular}{lr|cccccccc}
\toprule    
\multirow{2}{*}{Model} & \multirow{2}{*}{Size} & \multicolumn{4}{c}{High} & \multicolumn{2}{c}{Mid} & \multicolumn{2}{c}{Low} \\
\cmidrule(lr){3-6} \cmidrule(lr){7-8} \cmidrule(lr){9-10}
& & \texttt{en} & \texttt{fr} & \texttt{de} & \texttt{es} & \texttt{nl} & \texttt{it} & \texttt{cs} & \texttt{hu} \\
\midrule
SeamlessM4T-medium  & 1.2B  &  8.2 & 11.8  &  14.0 & 8.8  &  17.2    & 22.8 &  11.0  & 14.1\\
Canary              & 1.0B  &  6.0   & 9.2  &  10.7 & 7.0  &  -    & - &  -    & -\\
\midrule
Whisper-large       &1.5B   & 8.1 & 10.5 & 15.2 & 8.5 & 17.6  & 22.9 &  17.7    & 18.4 \\
Whisper-large-v2    &1.5B   & 7.9 & 10.4 & 13.1 & 7.9 & 15.8  & 23.2 &  14.3    & 18.3 \\
\midrule
Whisper-large-v3    &1.5B  & 9.7 & 10.4 & 19.7 & 10.6 & 24.9  & 32.3 &  12.4   & 16.3\\
~~~~+ Real-only (15K Hrs)    & -    & 6.0 & 8.9  & 9.6  & 7.0  & 12.5  & 21.7 &  9.5    & 11.7\\
~~~~+ Speech BT (500K Hrs)    & -    & 5.6 & 8.4  & 9.0  & 7.4  & 12.5  & 19.8 &  7.6    & 8.3\\
\bottomrule
\end{tabular}
\caption{Performance comparison across languages on Voxpopuli.}
\label{tab:whisper-vox}
\end{table*}

\begin{table*}
\centering
\begin{tabular}{lr|cccccccccc}
\toprule    
\multirow{2}{*}{Model} & \multirow{2}{*}{Size} & \multicolumn{5}{c}{High} & \multicolumn{2}{c}{Mid} & \multicolumn{3}{c}{Low} \\
\cmidrule(lr){3-7} \cmidrule(lr){8-9} \cmidrule(lr){10-12}
& & \texttt{en} & \texttt{fr} & \texttt{de} & \texttt{es} & \texttt{zh} & \texttt{nl} & \texttt{it} & \texttt{cs} & \texttt{hu} & \texttt{vi} \\
\midrule
SeamlessM4T-medium &1.2B  &  11.3   & 14.5  &  12.1 & 9.8  &  18.7  & 15.2 &  10.4  & 14.4 & 34.8 & 24.1\\
Canary             &1.0B  &  8.6   & 6.9  &  5.1 & 4.4  &  -    & - &  -    & - & - & -\\
\midrule
Whisper-large      &1.5B  & 12.2 & 15.0 & 8.9& 7.6& 17.3  & 8.1 &  10.1  & 19.9 & 23.8 & 24.5\\
Whisper-large-v2   &1.5B  & 11.7 & 13.7 & 7.8& 6.9& 16.9  & 6.9 &  9.3  & 16.5 & 20.3 & 22.8\\
\midrule
Whisper-large-v3            &1.5B  & 10.7 & 11.8 & 6.5& 5.5& 16.1  & 4.9 &  6.9 & 10.9 & 15.3 & 20.5 \\
~~~~+ Real-only (15K Hrs)  & -     & 9.7  & 8.7  & 5.9& 4.4& 14.3  & 4.3 &  5.5 & 9.2  & 11.4  & 16.9 \\
~~~~+ Speech BT (500K Hrs)  & -    & 8.8  & 7.3  & 5.0& 4.2& 13.6  & 3.7 &  4.9 & 5.2  & 6.0  & 13.6 \\
\bottomrule
\end{tabular}
\caption{Performance comparison across languages on Common Voice 16.}
\label{tab:whisper-cv}
\end{table*}

\end{document}